\documentclass[preprint,12pt]{RAS_template/elsarticle}

\usepackage{amssymb}
\usepackage{amsmath}

\usepackage{amsmath} 
\usepackage{amssymb}  
\usepackage{graphicx}
\usepackage{subcaption}
\usepackage{multirow}
\usepackage{float}
\usepackage[T1]{fontenc}
\usepackage[svgnames]{xcolor}

\DeclareMathAlphabet\mathbfcal{OMS}{cmsy}{b}{n}

\newcommand{\pos}{p}
\newcommand{\Stif}{\mathbf{K}} 

\newcommand{\Damp}{\mathbf{D}}
\newcommand{\Jacobi}{\mathbf{J}}

\newcommand{\jp}{\mathbf{q}}
\newcommand{\jv}{\mathbf{\dot{q}}}
\newcommand{\tq}{\boldsymbol{\tau}}
\newcommand{\wrench}{\mathbf{w}}

\newcommand{\pow}{P}

\newcommand{\powtankLOW}{\underline{\dot{E}}_{\text{T}}}
\newcommand{\entank}{E_{\text{T}}}
\newcommand{\entankUP}{\overline{E}_{\text{T}}}
\newcommand{\entankLOW}{\underline{E}_{\text{T}}}

\newcommand{\REAL}{\mathbb{R}}

\newcommand{\statespace}{\mathcal{S}}
\newcommand{\actionspace}{\mathcal{A}}
\newcommand{\rewardfunc}{R}
\newcommand{\rlconstraint}{\mathcal{C}}
\newcommand{\statetransprob}{P}

\newcommand{\storefunc}{S}

\usepackage{adjustbox}

\usepackage[inline]{enumitem}
 \usepackage{overpic}
 \usepackage[font=small,labelfont=bf]{caption}

\usepackage{pifont}
\usepackage{soul} 
\usepackage[colorlinks=true, linkcolor=blue, citecolor=red, urlcolor=magenta]{hyperref}  

\newcommand{\highlight}[1]{ #1}
\newcommand{\highlightnew}[1]{{\color{DarkGreen} #1}}

\newcommand{\cmark}{{\color{RoyalBlue} \ding{51}}}%
\newcommand{\xmark}{{\color{OrangeRed} \ding{55}}}%

\journal{Nuclear Physics B}

\begin{document}

\begin{frontmatter}



\title{Passivity-Centric Safe Reinforcement Learning \\for Contact-Rich Robotic Tasks}


\author[hri2,drim]{Heng Zhang} 
\ead{heng.zhang@iit.it}
\author[hri2]{Gokhan Solak} 
\author[hri2]{Sebastian  Hjorth} 
\author[hri2]{Arash Ajoudani} 

\affiliation[hri2]{
        organization={HRI$^2$ Lab, Istituto Italiano di Tecnologia},
        city={Genoa},
        country={Italy}
        }
\affiliation[drim]{organization={Ph.D. program of national interest in Robotics and Intelligent Machines (DRIM) and University of Genova},
        city={Genoa},
        country={Italy}}

\begin{abstract}
Reinforcement learning (RL) has achieved remarkable success in various robotic tasks; however, its deployment in real-world scenarios, particularly in contact-rich environments, often overlooks critical safety and stability aspects.
Policies without passivity guarantees can result in system instability, posing risks to robots, their environments, and human operators. 
In this work, we investigate the limitations of traditional RL policies when deployed in contact-rich tasks and explore the combination of energy-based passive control with safe RL in both training and deployment to answer these challenges. 
Firstly, we reveal the discovery that standard RL policy does not satisfy stability in contact-rich scenarios. Secondly, we introduce a \textit{passivity-aware} RL policy training with energy-based constraints in our safe RL formulation.
Lastly, a passivity filter is exerted on the policy output for \textit{passivity-ensured} control during deployment.
We conduct comparative studies on a contact-rich robotic maze exploration task, evaluating the effects of learning passivity-aware policies and the importance of passivity-ensured control.
The experiments demonstrate that a passivity-agnostic RL policy easily violates energy constraints in deployment, even though it achieves high task completion in training. 
The results show that our proposed approach guarantees control stability through passivity filtering and improves the energy efficiency through passivity-aware training. A video of real-world experiments is available as supplementary material.
We also release the checkpoint model and offline data for pre-training at \href{https://huggingface.co/Anonymous998/passiveRL/tree/main}{Hugging Face}.
\end{abstract}




\end{frontmatter}



\section{INTRODUCTION} \label{sec:intro} 
In recent years, RL has earned increasing attention and success in addressing complex decision-making and control problems, especially in robotic applications \cite{tang2024deep}. From manipulation tasks to autonomous navigation, RL offers the potential to achieve unprecedented performance by learning optimal control policies. However, while RL excels in simulated environments, deploying these policies in real-world robotic systems remains a significant challenge due to the safety concerns \cite{brunke2022safe}. 
This is particularly important in contact-rich tasks.
Contact-rich tasks require the robots to interact with their environment safely, particularly in terms of contact force regulation, which is necessary to avoid damaging the environment, the robot, or the objects involved in the task. 
Although, there has been a significant effort to answer the contact-rich safety problem, the role of the control stability is often neglected within RL.  

Stability is crucial for maintaining predictable and controllable robot behavior, especially when interacting with complex and unstructured surroundings~\highlight{\cite{kronander2016stability}}\cite{10806834} which can not be ignored in real-world robotic deployment~\cite{tang2024deep}. 
For instance, an RL policy trained without considering system stability might generate control commands that cause high-frequency vibrations or unbounded energy accumulation.
Such behaviors can damage the robot, compromise its performance, or pose safety risks to humans. 
Hence, ensuring that RL policies are inherently stable and safe 
is not only a theoretical concern but a practical necessity for robotics.
Despite these risks, a substantial portion of RL research focuses primarily on maximizing task success rates or improving learning efficiency, with limited attention given to the stability of policy. 
This gap presents a significant barrier to real-world deployment, where stability and safety are not only desirable but essential~\cite{bharadhwaj2022auditing}.  

\begin{figure}[tbh]
    \centering
    \includegraphics[width=0.65\linewidth]{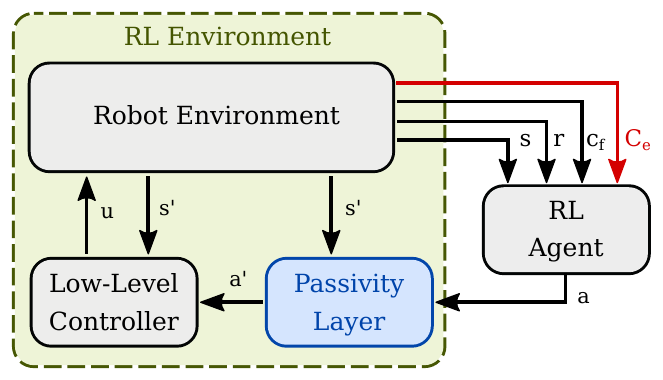}
    \caption{\textit{Passivity-ensured} Safe RL: The proposed method (blue and red colored elements highlight our contributions) adds both \textit{energy-based constraints} $C_e$ to enable learning the passivity-awareness and an active \textit{passivity-filtering layer} between the RL agent and the low-level controller for guaranteed passive control. The signals are the state $s$, reward $r$, action $a$, force constraint $c_f$ ,energy-based constraints $C_e$, and control command $u$.}
    \label{fig:ourmethod}
\end{figure}

Passivity-based control (PBC) is a fundamental method to prove the stability of physical interactions with the environment \cite{stefano2015earbook}. It is even more critical for kinematically redundant robots, as internal motions within the null space of the task Jacobian can result in an unintended increase in system energy, potentially leading to ~\cite{dietrich2016passive}.
Moreover, it is vital to monitor and restrict the amount of energy allowed to be stored and injected by the controller into the robot to ensure the passivity and safety of the system~\cite{lachner_shaping_2022},\highlight{\cite{raiola2018development,schwarz2022variable}}. Energy tanks are widely used to achieve passivity-based control by restricting the amount of energy allowed to be stored in the system~\cite{michel_safety-aware_2022}\highlight{\cite{beber2024passive,tveter2025passive}}. However, as mentioned above and pointed out in \cite{shahriari2019power}\highlight{\cite{tian2025passivity}}, the bounding of the energy alone is not sufficient to ensure the passivity of the system, but also the energy flow between the controller and robot has to be limited. We follow the recent works~\cite{michel_safety-aware_2022,michel_novel_2024,  SebastianICRA2023} with an augmented energy tank approach that also restricts the energy flow.
Although energy tank-based passivity is studied extensively in the classical control literature, \highlight{most of the research mentioned above focuses on the design of the energy tank and its parameters, while few works have investigated learning-based passive system design. 
Especially its inclusion in RL-based control learning is not sufficiently explored. Although some works~\highlight{\cite{9380915,jin2024learning}} leverage imitation learning method to obtain a stable VIC, labor-collected demonstrations are needed.} 

The inclusion of PBC in an RL system constitutes a significant gap in the RL literature.
In this work, we aim to explore different modes of combination between RL and PBC, and study their implications. 
We term the standard RL approaches that do not include any passivity treatment as \textit{passivity-agnostic} RL (Fig.~\ref{fig:passivity-agnostic}). 
Then, we distinguish between the \textit{passivity-filtered} approach, that imposes a passivity filter at the RL output;  and the \textit{passivity-aware} approach which aims to learn the passivity in the RL policy. 
In the former, the RL policy is trained traditionally, however, the filter analytically calculates the system energy and modifies the RL actions to stay passive in deployment. While it guarantees the passivity of the control, the RL agent remains unaware of the passivity. 
The latter introduces the energy awareness in the policy, however, it cannot guarantee the passivity as it is a black-box policy.
Subsequently, we propose the \textit{passivity-ensured} RL (Fig.~\ref{fig:ourmethod}) that includes both passivity guarantee and energy efficiency by combining the two approaches, as summarized in Table~\ref{tab:passivity-modes}. 

\textit{Passivity-aware} learning offers the possibility to obtain policies that solve tasks while providing benefits in terms of energy economy and long-term performance. Additionally, for systems with complex, nonlinear dynamics where analytical PBC design becomes intractable, learning provides a practical alternative that can discover passive policies without requiring complete system modeling.
Works such as~\cite{hathaway_learning_2023} and \cite{sacerdoti2024reinforcement} use RL to enhance designed PBC systems. \cite{hathaway_learning_2023} uses RL to learn the parameters of the control system. \cite{sacerdoti2024reinforcement} uses RL to dynamically adapt the size of an energy tank for changing tasks. 
To the best of our knowledge, only \cite{zanella2024learning} employs the classical energy-tank based passivity approach during RL training by filtering the RL output and terminating the training episode when the energy tank is depleted.
Although they train the system with an active filter, the RL agent is expected to learn the passivity only indirectly. 
\begin{table}
\renewcommand{\arraystretch}{1.4}
\caption{Method comparison}
\begin{adjustbox}{width=0.8\columnwidth,center}
\centering
\begin{tabular}{l||c|c|c|c}
  & \begin{tabular}[c]{@{}l@{}}passivity-\\ agnostic\end{tabular} & \begin{tabular}[c]{@{}l@{}}passivity-\\ filtered\end{tabular} & \begin{tabular}[c]{@{}l@{}}passivity-\\ aware\end{tabular} & \begin{tabular}[c]{@{}l@{}}passivity-\\ ensured\end{tabular} \\
  \hline \hline
\begin{tabular}[c]{@{}l@{}}passivity\\ guarantee\end{tabular} & \large{\xmark}    & \large{\cmark}  & \large{\xmark}   & \large{\cmark}                                                            \\
 \hline
\begin{tabular}[c]{@{}l@{}}energy\\ efficiency\end{tabular}   & \large{\xmark}  & \large{\xmark}   & \large{\cmark}  & \large{\cmark}                                       
\end{tabular}
    \end{adjustbox}
\label{tab:passivity-modes}
\vspace{-4.5mm}
\end{table} 

We advance the prior work 
by explicitly defining passivity-related constraints as part of the RL problem formulation. 
Furthermore, as stated in \cite{shahriari2019power}, solely limiting the total amount of energy is insufficient to ensure the passivity of the system. Hence, we augment the energy tank with a flow limit, unlike \cite{zanella2024learning}.
The main impact of this work lies in a systematic comparison of different approaches to combine PBC and RL.
Furthermore, we formulate our \textit{passive-aware} learning approach on a state-of-the-art safe RL framework to explicitly answer the contact-rich safety and passivity problems together. 
The safe RL concept has emerged to answer the safety concerns about the RL policies, which specify the safety requirements usually as constraints and aim to satisfy them while also solving the task~\cite{brunke2022safe}.
We adopt a recent safe RL method that combines variable impedance control (VIC) and safety critic frameworks for contact-rich tasks~\cite{HengRAL10517611}. 
Leveraging this safe RL framework, we take a step towards a passivity-ensured safe RL policy for robotic contact-rich tasks that achieves dual safe and stable performance. 

Lastly, we compare the four approaches of \textit{passivity-agnostic}, \textit{passivity-filtered}, \textit{passivity-aware} and \textit{passivity-ensured} RL extensively on a contact-rich interaction problem. 
The maze exploration task requires the robot to find the exit of a path by blindly touching the walls. 
Being contact-rich indicates a long exposure to physical contact. In fact, in this task, the contact is not only inevitable, but also fundamental to solve the task as the vision modality is not available. 
The robot has to regulate its interaction forces to solve this task effectively, benefiting from a VIC as shown on our previous work \cite{HengRAL10517611}.
Our experimental study investigates the effect of passivity-based formalisms on the training/learning phase of the policy.
Moreover, the performance in terms of task completion and energy expenditure of the different policies is compared to each other during deployment in simulation and the real world. The results confirm the advantages of the proposed approach in both guaranteeing the passivity and improving the energy usage while also safely solving a contact-rich task.

To summarize, the novel contributions of the manuscript are the following:
\begin{enumerate}
\item  Extensive comparison of different RL-based passive control learning approaches (Table~\ref{tab:passivity-modes}), and their effects on both training and deployment phases, evaluated from safety, stability, energy efficiency and task completion perspectives. 

\item Design and evaluation of the augmented energy tank-based passivity formulation in RL, with varying energy tank and energy flow constraints. 

\item Introduction of passivity-ensured safe RL framework that simultaneously answers both control passivity and contact-rich safety problems.

\end{enumerate}
\section{Passive Safe Reinforcement Learning} \label{sec:method}
The presented approach draws upon passive control and safe reinforcement learning fields to learn safe and stable policies for solving contact-rich tasks. 
As previously mentioned, it uses a passivity-based approach to observe and limit the energy stored in the system such that it can both:
\begin{enumerate*}
    \item impose constraints on the safe RL to train passivity-aware policies (Sec.~\ref{subsection:passivity-aware-rl}), and
    \item achieve passivity-ensured control by a passivity layer running on the policy output (Sec.~\ref{subsection:passivity-ensured-control}).
\end{enumerate*}
The former approach by itself does not guarantee passivity due to the incompleteness of data and the stochasticity of learning methods. The latter approach provides this guarantee, but can significantly reduce task performance by disrupting the task-oriented policy behavior. 
We combine both approaches to achieve good task performance while ensuring stability. 
A diagram of our proposed method is shown in Fig.~\ref{fig:ourmethod}.

\subsection{Safe RL for contact-rich tasks} \label{subsection:SRL4contact}
\begin{figure}[tbh]
    \centering
    \includegraphics[width=0.5\linewidth]{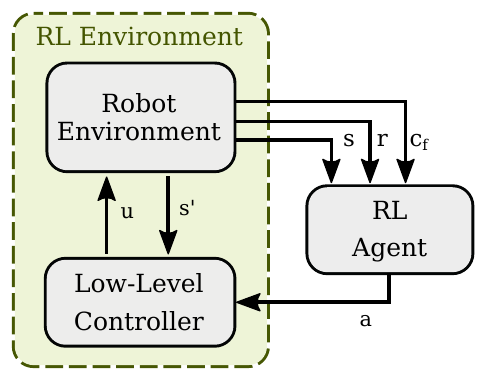}
    \caption{\textit{Passivity-agnostic} Safe RL: the baseline RL framework that we extend. 
    The baseline includes the force-based safety constraints $c_f$ in addition to the usual state $s$, action $a$ and reward $r$ observations of an RL framework.}
    \label{fig:passivity-agnostic}
\end{figure}
This section briefly introduces the contact-rich safe RL framework that forms the basis of our proposed method. 
The baseline safe RL framework (\textit{passivity-agnostic} in our terminology) is shown in Fig.~\ref{fig:passivity-agnostic}. 
Our RL agent consists of different components such as the \textit{safety critic} model and \textit{recovery policy}, in addition to the classic RL task policy.
The \textit{safety critic} is a separate network that predicts the potential risk of state-action pairs \cite{bharadhwaj2020conservative}. 
In the case of high risk, a \textit{recovery policy} is activated to escape from risky areas~\cite{thananjeyan2021recovery}.
The \textit{safety critic} model and \textit{recovery policy} are trained with the pre-collected offline data focusing only on the constraint violations. 
Then, all components including \textit{task policy} are trained online with the maze exploration task. 
The policy output as action is given to the VIC, which controls the robot to achieve the desired actions. 
We use a variable impedance action-space for its advantages in physical interaction tasks \cite{HengRAL10517611,martin2019variable}. 

Safety can be formulated by introducing constraints to the RL policy in the form of a constrained Markov decision process (CMDP) \cite{altman1995constrained}.
A CMDP is expressed as a tuple $(\statespace, \actionspace,\rewardfunc,\statetransprob, \gamma, \mu, \rlconstraint)$, where $\statespace$ is the state space, $ \actionspace$ is the action space, $\rewardfunc: \statespace \times \actionspace \rightarrow \REAL$ is the reward function, $\statetransprob$ is the state transition probability, $\gamma \in(0,1)$ is the reward discount factor, $\mu$ is the starting state distribution and $\rlconstraint=\left\{\left(c_i: \mathcal{S} \rightarrow\{0,1\}, \chi_i \in \mathbb{R}\right) \mid i \in \mathbb{Z}\right\}$ denotes the constraints that the agent must satisfy\highlight{, where $c_i(s)=1$ if the agent violates the constraint $i$, and $\chi_i$ is the allowed number of violations of that constraint. In our case, $\chi_i{=}0$, so that any constraint violation triggers termination}.
For this specific formalism, we define the state space, action space, reward function, and safety constraints as below. All vectors are in the world reference frame unless specified.
\begin{itemize}
    \item {State Space $\statespace$}. The state space is designed as A 6-dimensional vector for \textit{safety critic} network and \textit{recovery policy} including the wrench $[f_x, f_y, f_z, \highlight{t_x, t_y, t_z]}\in se^*(3)$ measured from an F/T sensor. Additionally, a 9-dimensional state vector is used \highlight{for the \textit{task policy}}, which extends this representation by incorporating the end-effector position [$p_x, p_y, p_z$]$\in\REAL^3$.
    
    \item {Action Space $\actionspace$}. Based on ~\cite{HengRAL10517611}, we design a 4-dimensional vector as action space, consisting of two end-effector displacement values $[\Delta \pos_x, \Delta \pos_y]\in\REAL^2$ and 2 stiffness parameters $[k_1, k_2]$. The action dimensions are bounded as listed in Table~\ref{tab:ctrlVar}. 
    We keep the position $\Delta \pos_z$ and stiffness $k_z\in\REAL$ in the z-axis constant. The stiffness values are defined w.r.t. the motion vector $\Delta \mathbf{\pos}$ with $k_1\in\REAL$ along the motion vector, $k_2\in\REAL$ orthogonal to the motion vector and parallel to the ground plane, and $k_z$ normal to the ground plane. 
    
    \item {Reward Function $\rewardfunc$}. The agent is penalized for the Euclidean distance between the current position of the end-effector and the goal position multiplied by a constant $r_{pos}$. High-force collisions are penalized with $r_{col}$. A bonus is given for reaching the goal $r_{goal}$ to increase convergence speed. The reward multipliers used in our experiments are presented in Table~\ref{tab:ctrlVar}.

    \item {Constraints $\rlconstraint$}. The force constraint $c_f$ requires the measured force magnitude to stay below a threshold. We also propose two new constraints for passive control, described later in Sec.~\ref{subsection:passivity-aware-rl}.
\end{itemize} 
In the following section, we describe how we apply the RL action with our low-level controller.
\subsection{Cartesian impedance control}\label{subsection:cart-imp-control}

The control scheme utilized in this work is a Cartesian VIC, for which the control torques of the impedance term $\tq^\top_{\text{IC}}\in\REAL^n$ in a quasi-static condition \highlight{consists of the torques generated by the cartesian spring $\tq^\top_{\text{K}}\in\REAL^n$ and the damping term $\tq^\top_{\text{D}}\in\REAL^n$,}
\begin{equation}\label{eq:tau_imp} 
    \tq^\top_{\text{IC}} = \tq^\top_{\text{K}} - \tq^\top_{\text{D}}.
\vspace{-0.1cm}
\end{equation} 
The torques generated by the cartesian spring $\tq^\top_{\text{K}}$ are generated by the elastic wrench ${\wrench^{\text{0},\text{EE}}_K}^\top\in se^{\ast}(3)$, which can be expressed as 
 \begin{equation} \label{eq:WrenchEE_K}
    {\wrench^{\text{0},\text{EE}}_K}^\top = \begin{bmatrix}
    	\Stif_t & \Stif_c \\ 
    	\Stif^\top_c & \Stif_r \\
    	\end{bmatrix} \Delta{\boldsymbol{\eta}}.
        \vspace{-0.05cm}
\end{equation}
Here, the infinitesimal displacement \highlight{of the twist $\boldsymbol{\eta}\in se(3)$} is described by $\Delta{\boldsymbol{\eta}}\in se(3)$. The positive definite matrix $\Stif_{c}\in\REAL^{3\times3}$ describes the coupling between the stiffness values of the rotational ($\Stif_{r}\in\REAL^{3\times3}$), and translational ($\Stif_{t}\in\REAL^{3\times3}$) axes.
In this work the values of translational stiffness matrix $\mathbf{K}_t$ are obtained by mapping the RL action stiffness matrix $\mathbf{K}_a = \operatorname{diag}(k_1, k_2, k_z)$ (Sec.~\ref{subsection:SRL4contact}) to the world frame as follows $\mathbf{K}_t = \mathbf{R}_p^\top \mathbf{K}_a \mathbf{R}_p$, given 
    \begin{equation*}
        \mathbf{R}_p {=}  \begin{bmatrix}
        \Delta \pos_x & -\Delta \pos_y  & 0\\
        \Delta \pos_y & \Delta \pos_x  & 0\\
    0 & 0  & 1\\
    \end{bmatrix} \in SO(3)
    .
    \end{equation*}
\highlight{The damping related torques $\tq_{\text{D}}$ are generated by Cartesian damping matrix $\Damp\in\REAL^{6\times6}$, 
which is computed directly from the stiffness matrix using a modal decomposition. We compute the eigen decomposition $\mathbf{K}_t = \mathbf{Q}\mathbf{\Lambda} \mathbf{Q}^\top$, where $\mathbf{\Lambda}$ contains the eigenvalues and $\mathbf{Q}$ contains the eigenvectors of $\mathbf{K}_t$. The damping matrix is then defined as $\mathbf{D}=2\mathbf{Q}(\xi\sqrt{\mathbf{\Lambda}})\mathbf{Q}^\top$, where $\xi$ is a scalar coefficient. This approach ensures that $\mathbf{D}$ remains symmetric and positive definite while being consistent with the variable stiffness updates.}

\highlight{We calculate the joint torques using the spatial Jacobian matrix $\Jacobi\in\REAL^{6\times n}$, rewriting \eqref{eq:tau_imp} as
\begin{equation}
    \tq^\top_{\text{IC}} = \Jacobi^\top\big({\wrench^{\text{0},\text{EE}}_K}^\top - \Damp\boldsymbol{\eta}\big).
\end{equation}}

In the following subsection, we describe the passivity-based approach for guaranteeing the stability of our controller.

\subsection{Passivity-based approach for variable stiffness control} \label{subsection:Passivity}
As mentioned in Section~\ref{sec:intro}, passivity is a necessity to facilitate physical interactions between a robot and its environment.
For an impedance-controlled robot, the energy stored in the system can be expressed as the sum of the storage functions of the controller and robot $\storefunc = \storefunc_{\text{Ctrl}}+\storefunc_{\text{Rob}}\in\REAL$ ~\cite{lachner_thesis_2022}.
It was shown in ~\cite{stefano2015earbook} that since $\storefunc_{\text{Rob}}$ is physically bounded, it is also passive. Therefore, the system is passive if and only if the controller described by $\storefunc_{\text{Ctrl}}$ is passive. Hence, it is a necessity to observe and bound the energy injected by the controller at the power port $(\tq_{\text{Ctrl}},\jv)$, which is expressed by
\begin{equation}
\dot{\storefunc}_{\text{Ctrl}} + \pow_{\text{IC}} = 0.
\end{equation}
Here, $\pow_{\text{IC}}$ is expressed as the sum of the energy flows resulting from the spring \highlight{$\pow_{\text{K}} = \wrench^{\text{0},\text{EE}}_K\boldsymbol{\eta}\in\REAL$ and damping $\pow_{\text{D}} = \boldsymbol{\eta}^\top\Damp\boldsymbol{\eta}\in\REAL$.} 
As mentioned previously in Sec.~\ref{sec:intro}, one can ensure the passivity of the overall system via the augmentation of the storage function $\storefunc_{\text{Ctrl}}$ by an energy tank $\entank\in\REAL$ that is bounded by the upper and lower limits $\entankUP$/$\entankLOW\in\REAL$.
Thus, the energy flow at the power port $(\tq_{\text{Ctrl}},\jv)$ can be rewritten as follows:
\begin{equation}
\dot{\storefunc}_{\text{Ctrl}} + \dot{\entank} + \pow_{\text{D}}\leq 0, 
\end{equation}
where the energy flow at the tank in this work is defined as $\dot{E}_{\text{T}} = -\pow_{K}$. 
The system passivity requires that the \textit{tank condition} $\entankLOW \leq \entank {+} \dot{\entank} \leq \entankUP$ to always hold. 

%
%
However, as pointed out in \cite{shahriari2019power}, satisfying the \textit{tank condition} is not enough to ensure the passivity of a system, but it is also a necessity to restrict the energy flow $\dot{\entank}$ between the controller and robot. 
This can be achieved by introducing a flow limit $\powtankLOW\in\REAL$, and ensuring that the \textit{flow condition} $\powtankLOW \leq \dot{E}_{T}$ always holds. 

In sections \ref{subsection:passivity-aware-rl} and \ref{subsection:passivity-ensured-control}, we go into more detail on how we incorporate \textit{energy tank} and \textit{energy flow} conditions into the learning and deployment phases, respectively.

\subsection{Passivity-aware safe RL} \label{subsection:passivity-aware-rl}

\begin{figure}[tbh]
    \centering
    \includegraphics[width=0.5\linewidth]{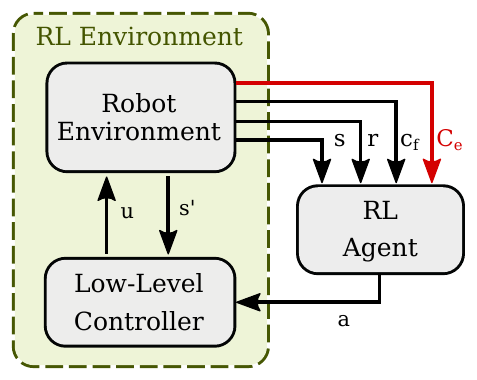}
    \caption{\textit{Passivity-aware} Safe RL: We introduce the energy constraints (${c_{eb},c_{ef}\in{C_e}}$) to enable the RL agent to learn passive behavior. }
    \label{fig:passivity-aware}
\end{figure}

This section describes how we add passivity awareness to the RL policy. 
As described in Section~\ref{subsection:SRL4contact}, the safe RL framework achieves safety through constraint definitions. 
We extend the CMDP formulation by including the passivity constraints so that the safety critic also learns the concept of passivity. 
As illustrated in Fig.~\ref{fig:passivity-aware}, this information is provided as the \textit{energy constraints} ${C_e} \subset \rlconstraint$, including a budget constraint ${c_{eb}\in{C_e}}$ and a flow constraint ${c_{ef}\in{C_e}}$, which correspond to the \textit{energy tank} and \textit{energy flow} conditions.
\highlight{Together with the force constraint $C_f$, they consist of the constraint $\rlconstraint \{ c_f,c_{eb}, c_{ef}\} $, formulating the CMDP where $c_i(s)=1$ iff threshold exceeded}.
The budget constraint ${c_{eb}}$ is introduced to make the agent aware of the amount of energy it has available for adapting the Cartesian stiffness $\Stif$ during the task. 
The constraint is triggered whenever $\entank + \dot{\entank} \leq \entankLOW$, which indicates that the lower bound of the energy tank is reached. 
\highlight{We encode the passivity information through the constraints \(C_e\), rather than augmenting the state space $\statespace$ with the tank variables, in order to directly shape the optimization objective while keeping the Markov state compact and focused on task-relevant quantities.}

Whereas the purpose of introducing the energy flow constraint ${c_{ef}}$ is to make the agent aware of how much energy it is allowed to take out of the energy tank at any given time for a specific task.
Therefore ${c_{ef}}$ is triggered whenever the draining energy flow of the energy tank exceeds the defined threshold $\dot{E}_{\text{T}} < \powtankLOW$. 
RL training episodes are terminated when a constraint happens. 
This makes it easy to end an episode with both of the energy constraints, hindering the training process. 
Thus, we experiment with different combinations of these constraints as detailed in Sec.~\ref{sec:eval}.
Defining the passivity information as constraints means that the training data includes this information.
This data is used to train the safety critic, which learns the risk related to passivity. 
Therefore, we expect the passivity-aware RL agent to learn to use the energy more economically, in means of cumulative energy spent and instant energy flow. 
However, learning a regression model does not guarantee that the risk is always predicted correctly. 
Thus, we recommend deploying these models with the added passivity layer for passivity-ensured control, as described in the following section.
\subsection{Passivity-ensured control}\label{subsection:passivity-ensured-control}
We employ a passivity layer at the output of the RL agent to guarantee the control passivity, as illustrated in Fig.~\ref{fig:ourmethod}. 
The passivity layer contains two filtering actions to satisfy the \textit{energy budget} and \textit{energy flow} conditions defined in Section~\ref{subsection:Passivity}.
We ensure the \textit{flow condition} \cite{shahriari2019power} by limiting the rate at which the controller can inject energy into the system through the introduction of the scaling variable
\begin{equation}\label{eq:alpha}
  \alpha = \begin{cases}
    \frac{\powtankLOW} {\dot{E}_{T}} & \text{if } \dot{E}_{T} < \powtankLOW \leq 0\\
    1 & \text{otherwise}.
 \end{cases}
\end{equation}
It is defined as the ratio between the maximal allowed energy flow $\powtankLOW$ and the originally calculated flow $\dot{E}_{T}$, resulting in a new formulation of the energy flow
\begin{equation}\label{eq:tank-condition}
    \dot{\entank}' = \begin{cases}
        \alpha\dot{\entank} & \text{if } \entankLOW \leq \entank + \dot{\entank} \leq \entankUP \\
        0& \text{otherwise}.
    \end{cases}
\end{equation}
Integrating the described energy tank dynamics for the VIC, the change of the Cartesian stiffness can be defined as
\begin{equation}
 \Stif(t) = 
    \begin{cases} 
    \Stif(t{-}1)        &     \text{if }    \entank + \dot{\entank}' \leq \entankLOW \\
    \Stif(t)    & \text{otherwise}. 
    \\
    \end{cases}
\end{equation}
In the event that the energy tank becomes depleted, $\Stif$ is hindered from increasing further, thereby keeping $\Stif$ constant. This results in a standard Cartesian impedance controller with constant gains. However, when $\entank>\entankUP$ becomes true through a decrease of $\Stif$, the change of stiffness is applied but without storing more energy in the tank.
In order to enforce the energy flow constraint, $\alpha$ is incorporated in the torques generated by the impedance control term:
\highlight{\begin{equation}\label{eq:tau_C_flow}
   {\tq}^\top_{\text{IC}}= \alpha{\Jacobi}^\top(\jp)\big({\wrench^{0,\text{EE}}_{K}}^\top + \Damp\boldsymbol{\eta}\big).
\end{equation}}
These filtering actions are necessary to guarantee passive behavior. 
However, reconsidering Fig.~\ref{fig:ourmethod}, the RL agent perceives the low-level control and the passivity layer as part of the environment, because these are not part of the RL-based decision-making.
Thus, any post-processing applied to the RL action decreases the transparency of the system, i.e., the outcome of the same action may be different depending on the filter variables. 
\highlightnew{If the passivity filter is applied during training, unsafe actions are corrected before affecting the environment, masking passivity violations from the agent and altering the action-to-outcome relationship. This creates a credit-assignment mismatch that limits the policy’s ability to learn intrinsically energy-aware behavior. Therefore, we use the passivity constraints to learn energy-aware policies during training, while reserving the passivity layer for guaranteeing stability during deployment.}

\section{Evaluation} \label{sec:eval}
We evaluate our method through the contact-rich maze exploration task. 
In the following evaluation experiments, we aim to study: 
\begin{enumerate*}
    \item how passivity constraints affect RL policy training and performance;
    \item how to guarantee both safe and stable performance while executing the task efficiently.
\end{enumerate*} 
To do so, we conduct three types of experiments. First, we train the agent with different energy budget and energy flow constraints, comparing them to the passivity-agnostic baseline (Sec.~\ref{sec:passivity-aware-exp}). Second, we deploy the learned models with the passivity layer (Sec.~\ref{sec:passivity-ensured-exp}), which guarantees control stability by limiting agent actions. Third, we validate our approach through real-world experiments (Sec.~\ref{sec:real_world}).
\begin{figure}[h!]
    \centering\includegraphics[trim=1cm 0.2cm 2cm 0.2cm,width=0.8\linewidth]{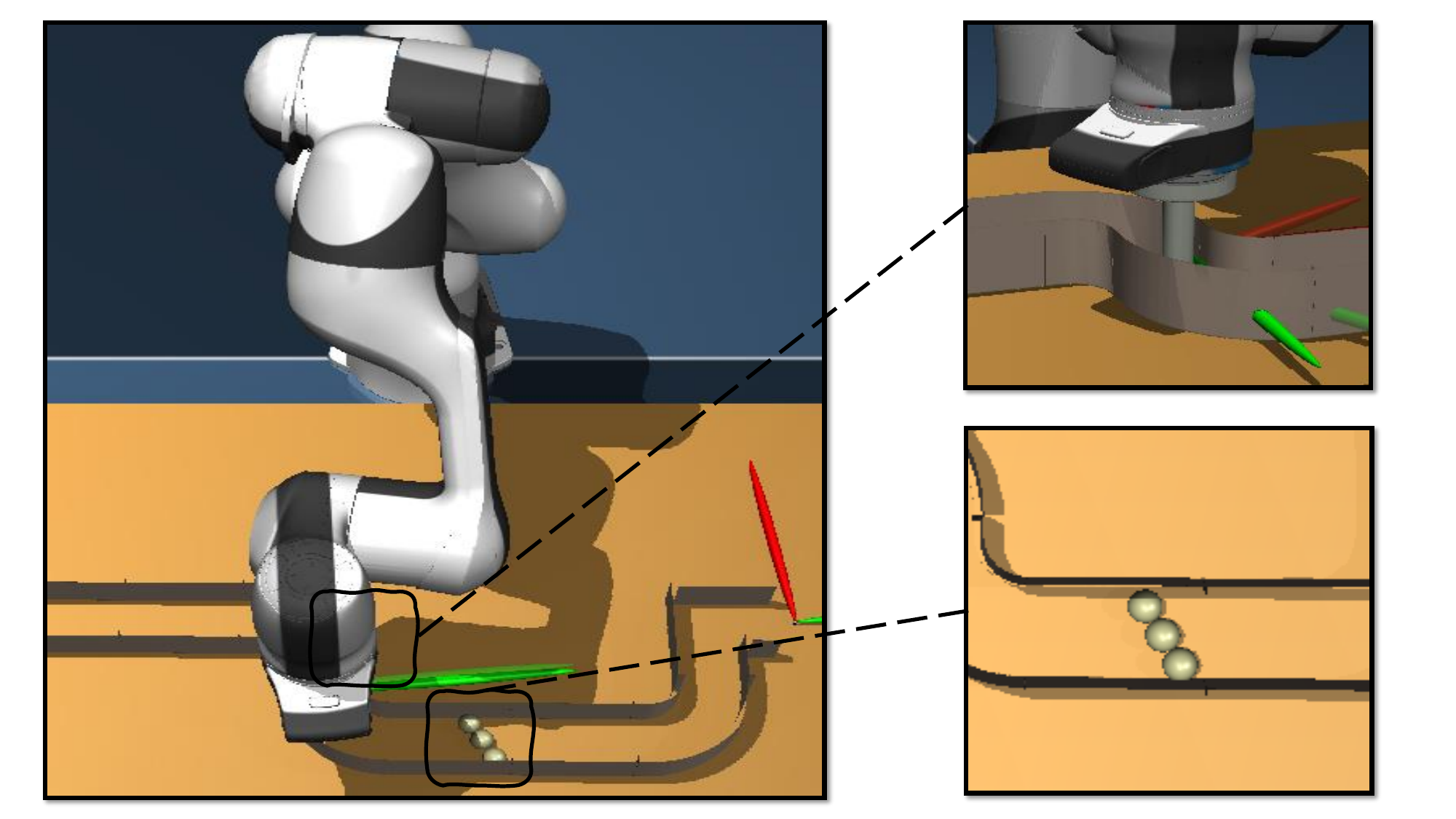}
    \caption{Experiments setup in MuJoCo. Right top: the robot explores the turn area with contact. Right bottom: three sphere obstacles. }
    \label{fig:exp_mujoco}
\end{figure}
\subsection{Experimental design}\label{sec:exp_des}
The experimental task is a maze exploration task where the robot has to explore an unknown maze blindly. 
Thus, it relies on the force sensing for finding its way out, making it a contact-rich task, i.e., the robot does not avoid the contact and it has to regulate the exerted wrenches for safety. During this process, the controller dynamically adjusts Cartesian stiffness and pose in response to disturbances, such as maze walls and movable obstacles represented by three simulated sphere balls in the middle of the maze. 

We train the policy in simulation using a Mujoco (version 2.3.3) maze environment, 
as shown in Fig. \ref{fig:exp_mujoco}.
First, 40,000 tuples with 1363 constraint violations were collected using the same method as in~\cite{HengRAL10517611} as offline data.  In comparison to \cite{HengRAL10517611}, we increased the force threshold to 40 N (same as in ~\cite{zhu2022contact}) to better focus on the passivity-based constraints $\entank$ and $\dot{E}_{\text{T}}$.
Afterwards, we trained policies through SAC~\cite{haarnoja2018soft} (see the basic hyper parameters in Tab.~\ref{tab:ctrlVar}) under different constraint configurations. Firstly, a passivity-agnostic policy where the agent focuses on safe task completion but is agnostic to system stability was trained. Followed by the training of the passivity-aware policies where the budget and energy flow constraints are included, by adding them to the CMDP. 
The policies were trained on the cluster ($1\times$ NVIDIA Tesla V100 16Gb GPU, $1\times$ Intel(R) Xeon(R) Silver 4210 CPU)
\footnote{We gratefully acknowledge the Data Science and Computation Facility and its Support Team at Fondazione Istituto Italiano di Tecnologia.}. 
For further reference, the following notation for the experiments is introduced. \texttt{Eb$\theta_b$} denotes experiments that are subjected to the energy tank constraint with size a of $\theta_b$${[J]}$. Furthermore, we introduce \texttt{Eb$\theta_b$-Ef$\theta_f$} for experiments that include both budget and flow constraints. Where \texttt{Ef} stands for energy flow and $\theta_f$${[W]}$ indicates the imposed energy flow limit (scaled by -10 to lose the decimal part).
For the real-world experiment setup please refer to \ref{sec:real_world}.
\begin{table}
\renewcommand{\arraystretch}{1.2}
\begin{adjustbox}{width=0.7\columnwidth,center}
\centering
    \begin{tabular}{l}
    \hline
    \hline
    \textbf{Reward multipliers}\\
        $r_{pos} = -400$,
        \space\space\space
        $r_{col} = -250$,
        \space\space\space
        $r_{goal} = 1000$\\
        \hline
        \highlight{\textbf{State space} }\\
           \highlight{ \textit{safety critic} and \textit{recovery policy}: $[f_x, f_y, f_z, t_x, t_y, t_z]\in se^*(3)$ }
            \\
           \highlight{ \textit{task policy}: $[f_x, f_y, f_z, t_x, t_y, t_z]\in se^*(3)$, $[p_x, p_y, p_z$]$\in\REAL^3$ }\\
        \hline
        \textbf{Action space} \\
            $\Delta p_x, \Delta p_y \in$  [-3, 3] cm,
            \space\space\space
            $k_1, k_2 \in$ [300, 1000] N/m \\
        \hline        
         \textbf{Control parameters}\\
         $k_z =$  750,
         \space\space\space
         $\mathbf{K}_r =\operatorname{diag}(100, 100, 0)$,\space\space\space $\mathbf{K}_c =\mathbf{0}$ N/m, \space\space\space \highlight{$\xi=0.707$}\\
         \hline
        \textbf{Training hyperparameters}\\
        learning rate: $3 \times 10^{-4}$, \space\space $\gamma$: 0.9, \space\space $\gamma_{safe}$: 0.85, \space\space $\epsilon_{risk}$:0.65 \\
    \hline
    \hline
    \end{tabular}
    \end{adjustbox}
    \caption{Experiment parameters}
\label{tab:ctrlVar}
\end{table}
\subsection{Passivity-aware policy training}\label{sec:passivity-aware-exp}
\begin{figure}[h!]
    \centering
    \includegraphics[width=0.7\linewidth]{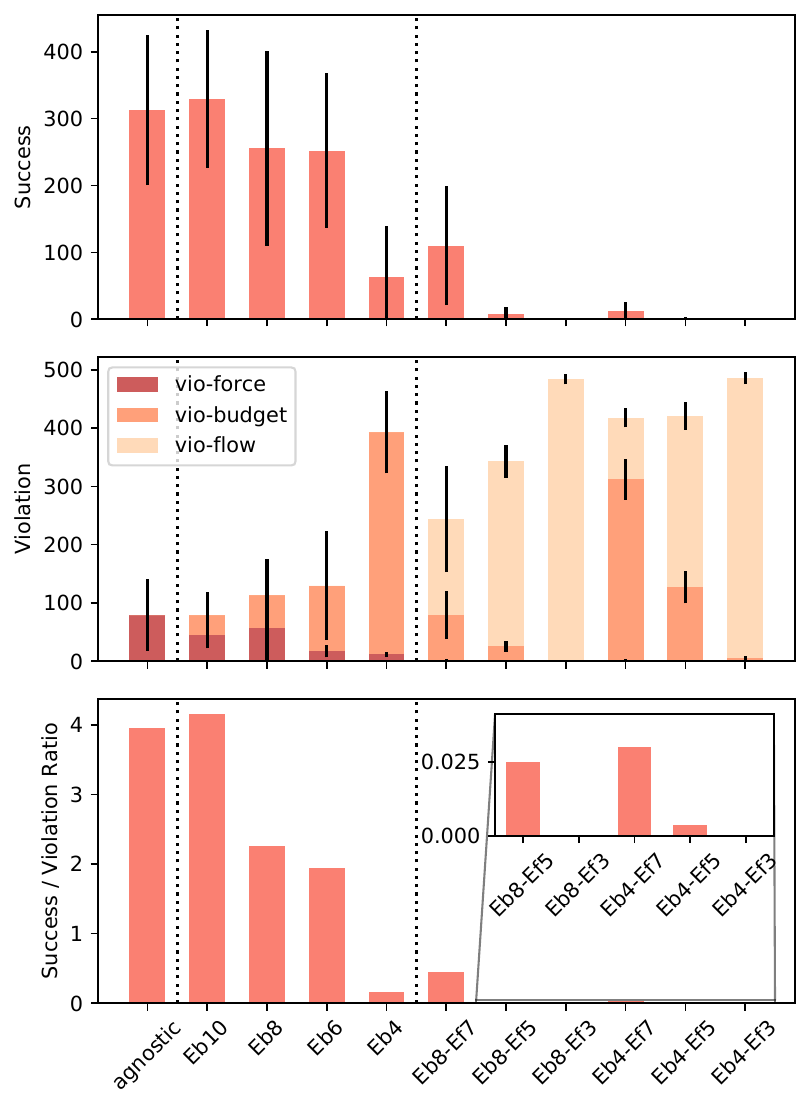}
    \caption{Passivity-aware policy training experiment results. We present the total numbers of successes and violations in the first two plots. The third plot shows the ratio of mean success to mean violation.
    }
    \label{fig:training-bar-results}
\end{figure}

As mentioned previously, this section focuses on how the passivity constraints affect the training of an RL policy, by comparing them to each other as well as to the \textit{passivity-agnostic} approach that is solely subjected to the contact-force constraint as shown in Fig.~\ref{fig:passivity-agnostic}.
\highlight{More specifically, the remainder focuses on energy budgets in the range of $4{-}10$ and flow limits from $-0.7$ to $-0.3$, identified through preliminary trials; outside this range, the constraints are either overly restrictive, hindering feasibility, or overly permissive, limiting their effect on learning passivity-aware policies.}
The results of the training experiments are summarized as means and standard deviations of 10 seeds in Fig.~\ref{fig:training-bar-results}. 

More specifically, it shows the amount of success in terms of task completion, the type and number of constraint violations which led to the task termination as well as the ratio between the amount of successfully completed runs and violations. 
 
Firstly, it can be seen that the task's success is directly correlated with the amount of constraints introduced to the CMDP. 
Secondly, the introduction of $c_{eb}$ has a direct effect on the amount of force violation, which overall decreases as the energy budget gets smaller. Furthermore, the introduction of $c_{ef}$ results in the force constraint no longer being violated.  The only outlier is the \texttt{Eb10}, which outperforms the agnostic run, even though it violates both the budget and force constraint. 
Thirdly, it can be seen that the introduction of $c_{ef}$ has a more aggressive effect on the task success than the introduction of only $c_{eb}$. This is due to the fact that the constraint on the flow hinders the agent from making aggressive changes in its stiffness, resulting in an overly conservative approach that prevents the agent from exploring the maze effectively. 
Hence, generally, it can be said that there is a consistent increase in constraint violations as the budget size decreases and harsher flow limits are applied. 

\begin{figure}[tbh]
    \centering
    \includegraphics[width=0.9\linewidth]{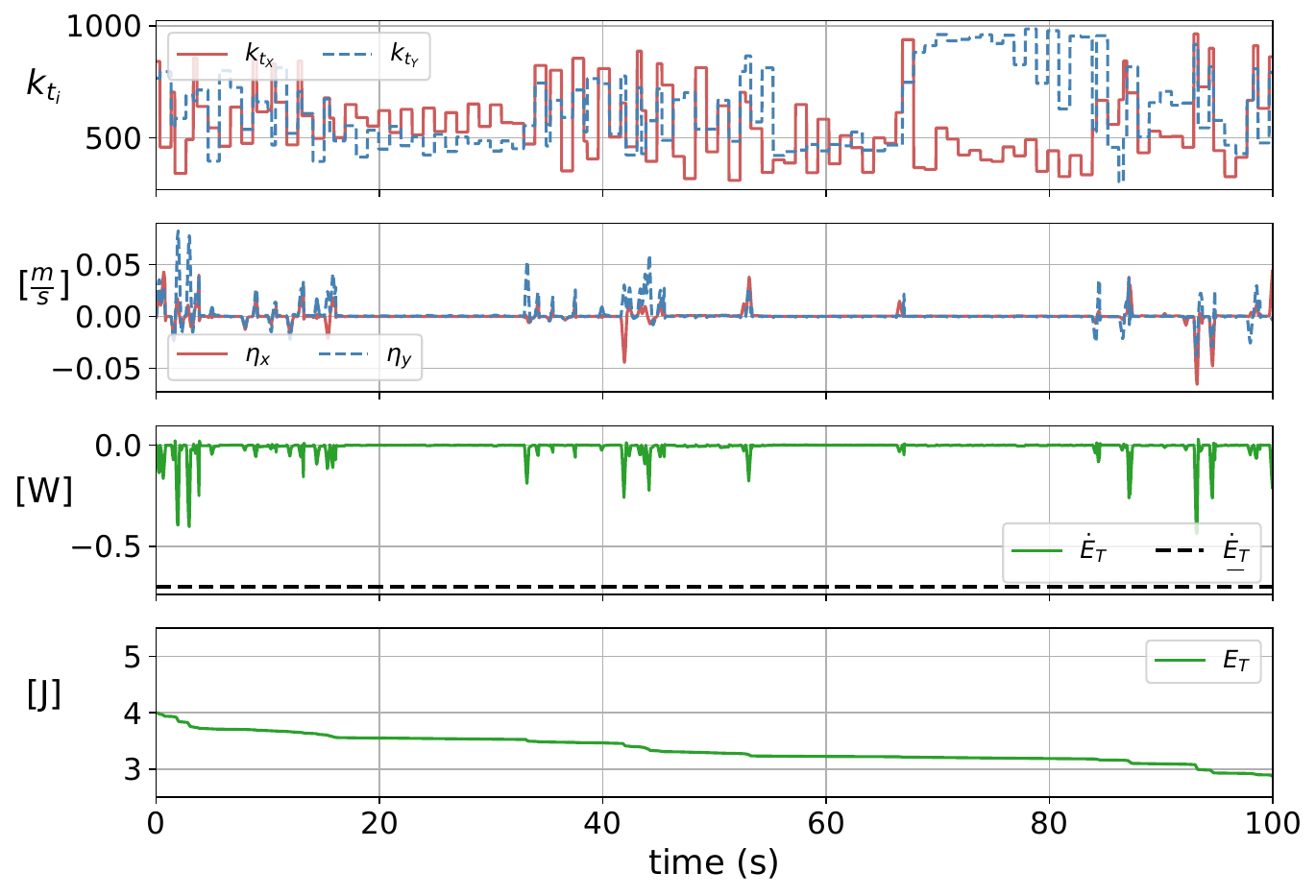}
    \caption{\highlightnew{These plots visualize the effect that the changes of the stiffnesses \highlight{$k_x$,$k_y$}$\in\Stif_\text{T}$ together with the Cartesian velocities of the end-effector $\eta_x,\eta_z\in\boldsymbol{\eta}$ have on the energy flow $\dot{\entank}$ and consequently on the energy stored in the energy tank $\entank$ during parts of the training.}} 
    \label{fig:stiffness}
\end{figure}
\begin{figure}[tbh]
    \centering
\includegraphics[width=0.8\linewidth, trim={0.3cm 0 0.25cm 0},clip]{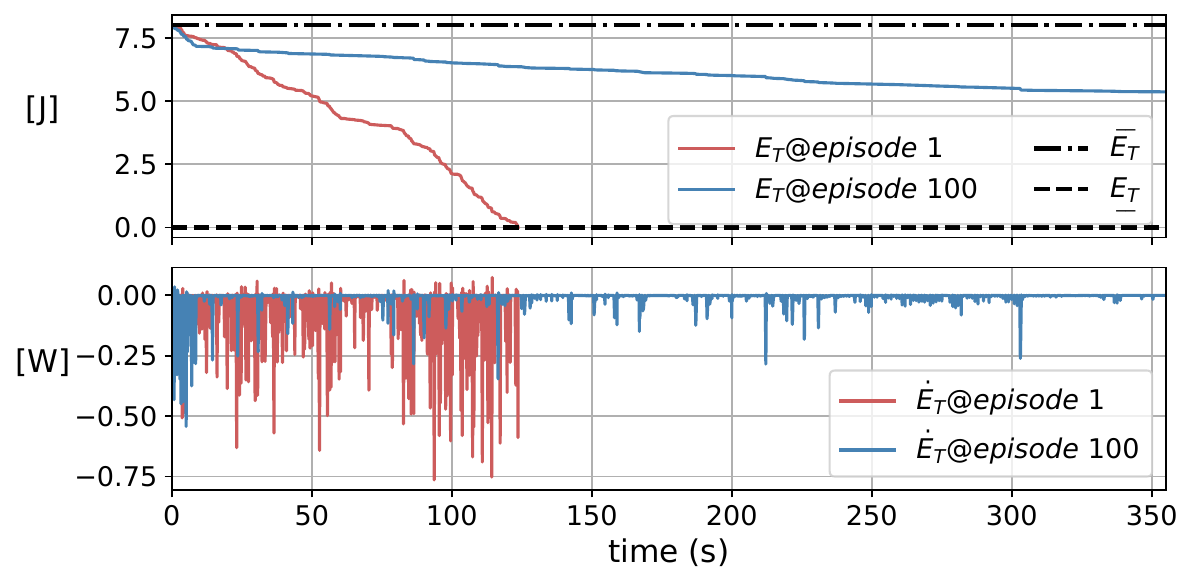}
    \vspace{-0.2cm}
    \caption{Plots illustrate the effect that the introduction of the \textit{energy tank constraint} has on the agents energy expenditure at different stages during training.}
    \label{fig:tank learning}
\end{figure}
%
During training the agent regulates the stiffness of the impedance controller (Sec.~\ref{subsection:cart-imp-control}). This has a direct effect on how the energy flow $\dot{\entank}$ and the energy tank budget $\entank$, which is visualized in Fig.~\ref{fig:stiffness}.
More specifically, it is shown that whenever the agent increases the stiffness of springs during its motion, it results in $\dot{\entank}<0$ as long as $\entankLOW\leq\entank<\entankUP$ holds as described in Sec.~\ref{subsection:Passivity}. 
Furthermore, it can be observed that the greater the increase in stiffness during a particular motion the greater effect it has on $\dot{\entank}$ and consequently on $\entank$. It is important to note that $\dot{\entank}$ does not solely depend on one specific spring stiffness, but on $\Stif$, the energy flow generated by a change of the stiffness of one spring can be offset by a change of the stiffness of another stiffness in the opposite direction. \highlightnew{Additionally, $\dot{\entank}$ is depended on the $\boldsymbol{\eta}$, meaning that that if the robot is stationary ($\boldsymbol{\eta}=0$) during the regulation of $\Stif$, no energy flow happens ($\dot{\entank} = 0$).}  
This behavior can be directly related to the definition of $\dot{\entank}$ (Sec. \ref{subsection:Passivity}).
During training, one can observe how the agent becomes more aware of how the change of $\Stif$ effects either $\dot{\entank}$ and/or $\entank$ depending on the constraints it is subjected to. 
The impact that $C_e$ has on passivity-aware policy training can be seen in Fig.~\ref{fig:tank learning}, showing the agent's energy expenditure at two different stages during training where the red graph representing the energy expenditure and the energy flow at the port of the energy tank in the first episode, and the blue graph represents the agent's energy-related behavior after 100 episodes.  When comparing both graphs, it is evident that the agent \highlight{becomes more conservative over the course of training under the proposed energy constraints, resulting in reduced and more evenly distributed energy expenditure. This figure is intended as a qualitative illustration of how exposure to the energy constraints $C_e$ leads the agent to adjust its energy usage over time, rather than as a standalone causal comparison to a passivity-agnostic baseline. A direct comparison between constrained and unconstrained policies is provided in Fig.~\ref{fig:deployment-energy-comparison}, where we contrast the passivity-aware and passivity-agnostic settings under identical task and deployment conditions.}

\begin{table}[th]
\renewcommand{\arraystretch}{1.1}
\caption{Results of model tests. Performance evaluation over 100 episodes for each of three randomly selected seeds. \highlight{“–” indicates that the corresponding constraint is not enforced or evaluated.}}
    \label{tab:model_test0}
    \centering
    \begin{tabular}{c|ccc|c}
        \hline
        \hline
    \multirow{2}{*}{\textbf{Run config.}}  & \multicolumn{3}{c|}{\textbf{Violations}} & \textbf{Successes}\\
      &Force & Tank& Flow & $\mu\pm\sigma$ \\
    \hline 
    agnostic &0 & \highlight{--}& \highlight{--} & $100\pm0$\\
    \hline 
     Eb4 &0 & 26& \highlight{--} & $91.3\pm3.9$\\
    \hline 
     Eb8  &2 & 0& \highlight{--} & $99.3\pm0.9$\\
    \hline 
      Eb8-Ef7 & 0 & 7& 0 & $62\pm40.1$\\
    \hline 
    \hline 
    \end{tabular}
\end{table}
After training, we selected the policies \highlight{of the best performing seeds} for \texttt{agnostic, Eb4, Eb8, Eb8-Ef7} to test the model in representative experiments. All policies are evaluated over 100 episodes under the same constraint settings used during training. \highlight{For the passivity-agnostic policy, this means that energy-related constraints are neither enforced nor evaluated.} In the testing results, shown in Table~\ref{tab:model_test0}, similar to the training results shown in Figure~\ref{fig:training-bar-results}, the agent achieves the highest success rate when it is agnostic to $C_e$  and can therefore operate more exploratory during its runs. Furthermore, looking at \texttt{Eb8} and \texttt{Eb4}, a more conservative energy tank size results in more tank violations. However, compared to the training results, the \texttt{Eb8-Ef7} run obtained fewer successes as it has harsher constraints during results.  Here, \texttt{Eb4} obtains high success, even though it had fewer successes than \texttt{Eb8-Ef7} in training. From this it can be concluded that \texttt{Eb4}  learned a better task policy, although it was delayed by its limited energy budget.


\subsection{Passivity-ensured deployment}\label{sec:passivity-ensured-exp}
In the deployment experiments, we evaluate the effect of the passivity layer on the trained models, by comparing the performances of the \textit{passivity-agnostic} baseline and various \textit{passivity-aware} approaches in combination with and without the passivity layer. We label the passivity-agnostic approach with the passivity layer as \textit{passivity-filtered} and the passivity-aware approaches with the layer as \textit{passivity-ensured}.
In these experiments, we do not retrain the models but impose the passivity layer on the output of the RL agent as in Fig.~\ref{fig:ourmethod}.  The passivity layer is applied with uniform parameter settings: an initial energy budget of $\entankUP{=}6$ and an energy flow limit of $\dot{\entankLOW}{=}-0.5$. 

\begin{figure}[tbh]
    \centering
\includegraphics[width=0.9\linewidth]{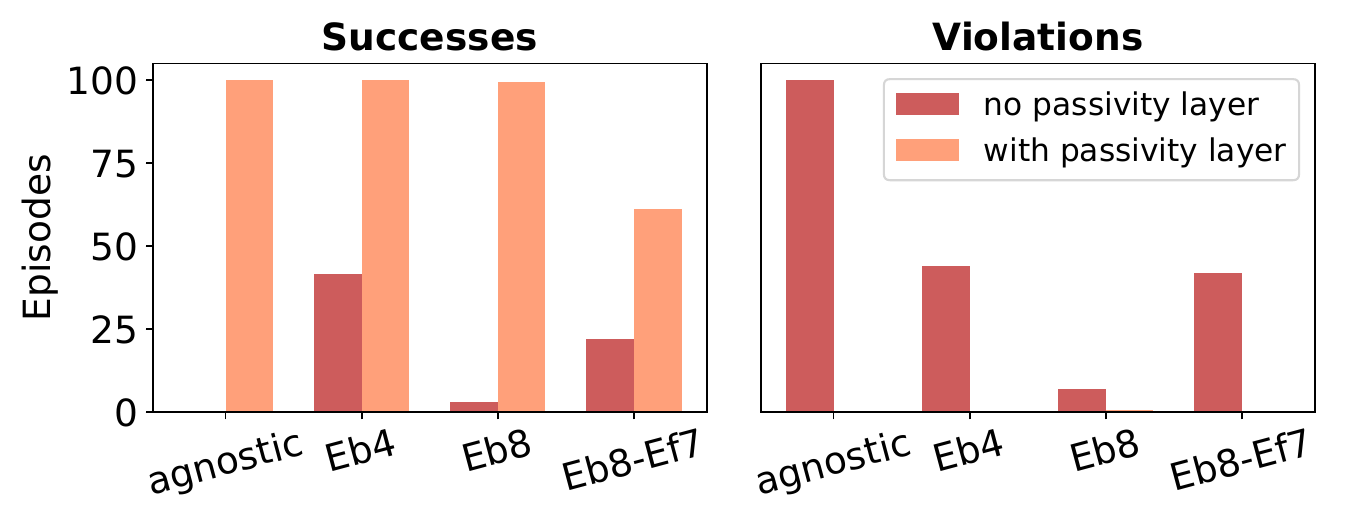}
    \vspace{-0.1cm}
    \caption{Policy performance in deployment experiments. The bars show how the models (\texttt{agnostic}, \texttt{Eb4}, \texttt{Eb8}, \texttt{Eb8-Ef7}) perform in terms of successes and violations with and without the passivity layer in simulation for 100 episodes, subjected to the same constraints $\entankUP{=}6$, $ \dot{\entankLOW}{=}-0.5$. 
    }
    \label{fig:deployment-bar-results}
\end{figure}
Three best-performing models of the previous section are selected and tested $100$ times each. 
We applied the same filter parameters $\theta_b{=}6$ and $\theta_f{=}0.5$ in all cases. 
The results in Fig.~\ref{fig:deployment-bar-results} showcase the task successes and violations. 
Overall, the results labeled as ``with passivity layer'' outperform the baseline labeled as ``no passivity layer''. These results reveal the risks of deploying models without passivity guarantees, which potentially leads to system instabilities and unsafe behaviors despite good performance in model tests shown in Table~\ref{tab:model_test0}. 
\begin{figure}[th]
\centering
    \begin{tabular}{c}
\begin{overpic}[trim=0.1cm 1.24cm 0.02cm 0.0cm,width=0.95\linewidth]{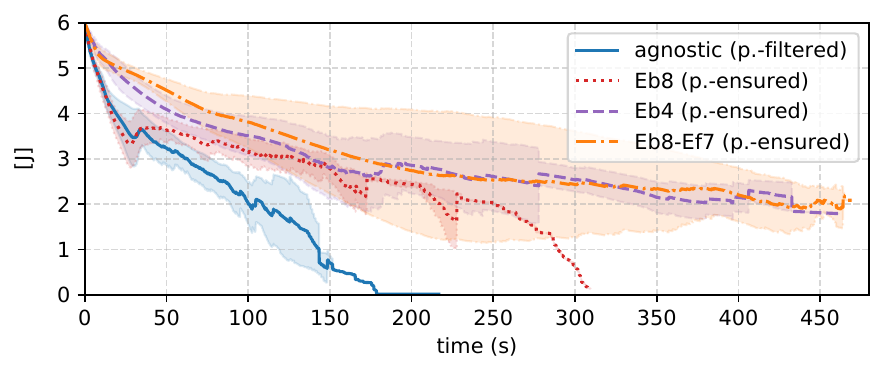}
\put(22,27){\footnotesize  \colorbox{white}{{\bfseries (a)} Energy budget progress}}
\end{overpic}
\\
\begin{overpic}[trim=0.0cm 0.40cm 0.00cm 0.0cm,width=0.95\linewidth]{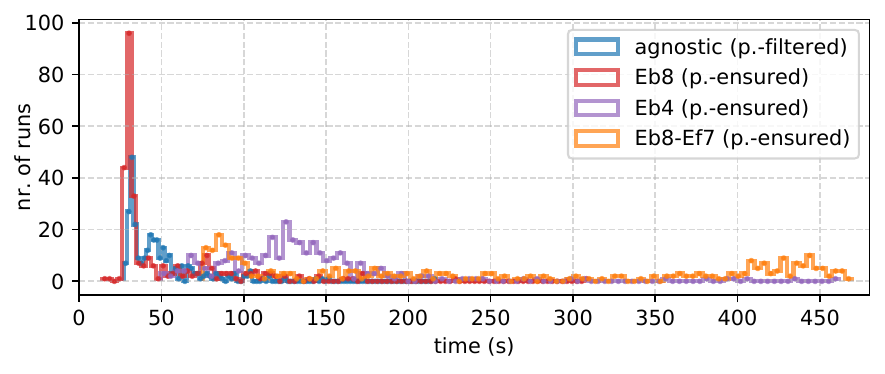} 
\put(22,33){\footnotesize  \colorbox{white}{{\bfseries (b)} Run duration histogram}}
\end{overpic}
    \end{tabular}

    \caption{\textbf{(a)} Energy budget progress during passivity-ensured deployment experiments. The curves indicate the mean of 300 runs (100 per seed) and the shades indicate the standard deviation. 
    \textbf{(b)} Histogram of the run times, i.e., the number of runs that finished in time duration intervals.
    }
    \label{fig:deployment-energy-comparison}
\end{figure}

Specifically: 
\begin{enumerate*}
   \item  The \texttt{agnostic} policy performs the worst when no passivity filter is applied, as it consistently violates the flow constraint. Although it achieves a 100\% success rate when evaluated without considering energy constraints shown in Table~\ref{tab:model_test0}, which highlights the necessity of incorporating passivity during training. 
   \item The success rate improves significantly when the passivity filter is applied, underscoring the dual benefits of passive-aware policies and the passivity layer, ensuring safer and more stable deployments.
   \item  Violations dropped significantly with the passivity layer, with only \texttt{Eb8} having two force violations that are almost not visible in Fig.~\ref{fig:deployment-bar-results}. Note that no energy constraint violations occurred, as the passivity layer ensures system passivity.
\end{enumerate*} 

We also present the energy usage of each method in Fig.~\ref{fig:deployment-energy-comparison}.a to further demonstrate the benefit of passivity-aware training. 
Even though all of \texttt{agnostic}, \texttt{Eb8} and \texttt{Eb4} show full success in the tests (Fig.~\ref{fig:deployment-bar-results}), analysis of the energy budget progress in these runs reveals that the \texttt{agnostic} model depletes its budget more rapidly. 
This rapid depletion reduces its likelihood of success in more complex task instances. On the other hand, \texttt{Eb4} demonstrates the best success and energy efficiency, despite its delayed learning shown in Fig.~\ref{fig:training-bar-results}. 
\highlight{In summary, the \texttt{agnostic} policy depletes the energy budget more rapidly due to its aggressive stiffness adaptation, which may reduce its robustness in longer or more complex tasks, while passivity-aware policies achieve a more favorable balance between speed and efficiency.}
Furthermore, the histogram in Fig.~\ref{fig:deployment-energy-comparison}.b shows number of runs distribution on completion time. \highlight{This figure shows that the \texttt{Eb8} and \texttt{agnostic} policies achieve comparable completion times, with \texttt{Eb8} being slightly faster on average despite its lower energy consumption.} \highlight{This indicates that the \texttt{agnostic} policy expends more energy per unit progress, rather than simply completing the task faster.} \highlight{In contrast, more conservative policies (\texttt{Eb4}, \texttt{Eb8-Ef7}) exhibit a trade-off between efficiency and completion time.}
%
\begin{figure}[th]
    \centering
    \begin{overpic}[trim=0.1cm 0.5cm 0.1cm 0.5cm,width=0.92\linewidth]{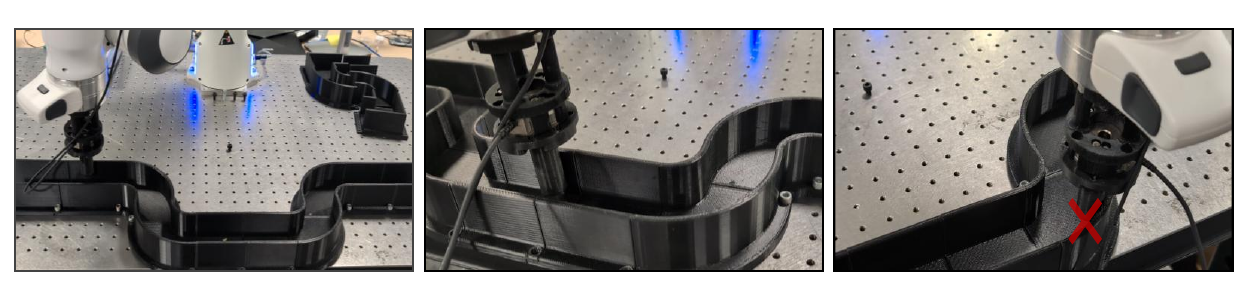} 
      \put(1, 15.5){\small \bfseries \colorbox{white}{a}}
      \put(34, 15.0){\small \bfseries \colorbox{white}{b}}
      \put(67, 15.5){\small \bfseries \colorbox{white}{c}}
    \end{overpic}
    \\
    \centering
    \begin{overpic}[width=0.95\linewidth]{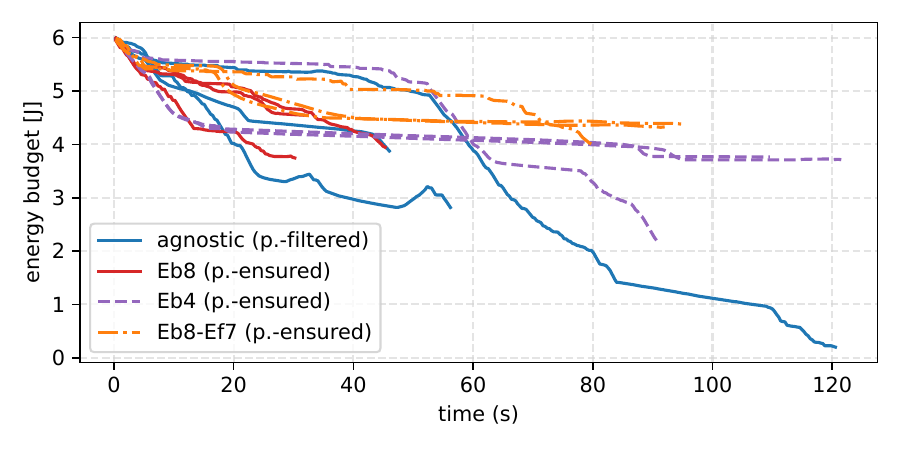}
        \put(88,41){\small \bfseries \colorbox{white}{d}}
    \end{overpic}
    
    \vspace{-0.3cm}
    \caption{Snapshots and result analysis in real-world experiments. \textbf{(a)} the overview of the maze exploration task. \textbf{(b)} closeup when the robot moves along with contact force. \textbf{(c)} failure case  
    agnostic but non-passive behavior at the turnaround area.
    \textbf{(d)} shows the energy budget consumption among different setups, demonstrating our method outperforms in terms of passivity and energy efficiency.
    }
    \label{fig:real-word exp}
\end{figure}
\subsection{Real-world experiments}\label{sec:real_world}
The trained models are deployed in the real world to complete our validation on a Franka Panda robot arm with a flange mounted at the end effector with an ATI45 FT sensor to perceive external force and torque as shown in Fig.~\ref{fig:real-word exp} upper, 
with our passivity-ensured impedance controller (frequency 1000Hz). 

We run the best models of each setup 3 times in the real world without fine-tuning. 
The robot can finish the task without violating force constraints and energy constraints in each experiment, which validates our method's effectiveness. 

However, \highlight{a failure case happened in the baseline (passivity-agnostic) model. }
\highlight{All passivity-ensured configurations (Eb4, Eb8 Eb8-Ef7) achieved a 100\% success rate over three trials each, with no constraint violations. In contrast, the passivity‑agnostic policy failed in one out of three trials, largely due to its more aggressive energy usage, applying excessively high force at the turning point. (Fig.~\ref{fig:real-word exp}.c).} 

Specifically, the Fig.~\ref{fig:real-word exp}.d shows all models' energy expenditure over time steps. First, we can observe that the \texttt{agnostic} consumes energy most aggressively although keeps passive due to the passivity layer, 
while passivity-aware models maintain lower energy consumption rates. Notably, \texttt{Eb4} achieves the most efficient energy usage among all models. This suggests that training with tighter energy constraints leads to more energy-efficient behaviors that transfer well to real-world deployment. 
The overall trend validates that our passivity-aware approach successfully learns energy-efficient policies that retain effectiveness in real-world settings.
The video is available as supplementary material. 

\subsection{Discussion}

\highlight{This section summarizes and discusses the performance of the different agents during training and deployment.
Firstly, the training results in Sec.~\ref{sec:passivity-aware-exp} show that the agent learns to be more conservative in its energy expenditure when subjected to the passivity-based constraints $C_e$ as directly observed in Fig.~\ref{fig:tank learning}.
Furthermore, Fig.~\ref{fig:training-bar-results} showcase that the inclusion of these constraints directly affects the success rate of task completion during training. This is because these constraints restrict the agent's exploration, which is crucial, especially at the beginning of the learning phase. However, although they learn the task later (Fig.~\ref{fig:training-bar-results}) constrained cases can still achieve high success rates in the model test (Table~\ref{tab:model_test0}), suggesting that the passivity-constrained cases would benefit from further training compared to the agnostic case. 
On the other hand,} overly restrictive constraints can prevent learning entirely if the task becomes physically infeasible.

\highlight{Importantly, the differences between the passivity-agnostic and constrained policies in Fig.~\ref{fig:training-bar-results} and \ref{fig:deployment-energy-comparison} indicate that the observed changes in energy usage are not merely a byproduct of task progress but arise from the explicit inclusion of $C_e$ during training.}


\highlight{Another important observation can be made in the deployment results shown in Figure~\ref{fig:deployment-bar-results}. It shows} that passivity cannot be guaranteed without a passivity layer, regardless of whether the agent was trained with passivity awareness  \highlight{or not. Furthermore, it showcases that while} the passivity layer improves task completion and reduces violations for all agents, those trained with energy constraints ($C_e$) demonstrate better energy efficiency.


The correlation between the energy usage trends in simulation (Fig.~\ref{fig:deployment-energy-comparison}.a) and real-world experiments (Fig.~\ref{fig:real-word exp}.d) demonstrates the successful transfer of learned energy-efficient behaviors. 
In both cases, the \texttt{agnostic} model shows the steepest decline in energy budget, while \texttt{Eb4} maintains the most conservative energy usage. This consistent pattern validates that the energy-aware policies learned in simulation maintain their characteristics when deployed on physical hardware. 
The real-world results also confirm that tighter energy constraints during training (\texttt{Eb4}) lead to more efficient policies that better manage the energy budget in both simulated and real environments. 
However, the absolute rate of energy consumption differs between simulation and reality due to modeling discrepancies and real-world friction effects.
%
%
\section{Conclusion and future work}
In this work, we investigated the safety and stability of the robotic system under multiple constraints for a contact-rich task, demonstrating that the traditional RL approach can not guarantee passivity. Furthermore, we combined an augmented energy tank-based passivity formalism in a safe RL framework. We employed the passivity constraints in both training and deployment of the RL model. The proposed method completed the task without any force violation while ensuring system passivity. Additionally, passivity-aware policies showed improved energy efficiency in deployment, even though they converged later in the training. A set of experiments both in simulation and in the real world on a 7-DoF Franka robot arm validated the effectiveness of the proposed method. We are planning to \highlight{develop more industrial applications to evaluate the method involving assembly operations that require higher interaction forces and} explore self-adaptive energy tank size and flow constraints in future work.

\section*{Acknowledgement}
This work was partially supported by the European Union Horizon Project TORNADO (GA 101189557).

 \bibliographystyle{elsarticle-num} 
 \bibliography{RAS_main}






\end{document}